# "Grip-that-there": An Investigation of Explicit and Implicit Task Allocation Techniques for Human-Robot Collaboration

KARTHIK MAHADEVAN

University of Toronto, karthikm@dgp.toronto.edu

MAURÍCIO SOUSA

University of Toronto, mauricio.sousa@utoronto.ca

ANTHONY TANG

University of Toronto, tonytang@utoronto.ca

TOVI GROSSMAN

University of Toronto, tovi@dgp.toronto.edu

In ad-hoc human-robot collaboration (HRC), humans and robots work on a task without pre-planning the robot's actions prior to execution; instead, task allocation occurs in real-time. However, prior research has largely focused on task allocations that are pre-planned - there has not been a comprehensive exploration or evaluation of techniques where task allocation is adjusted in real-time. Inspired by HCI research on territoriality and proxemics, we propose a design space of novel task allocation techniques including both explicit techniques, where the user maintains agency, and implicit techniques, where the efficiency of automation can be leveraged. The techniques were implemented and evaluated using a tabletop HRC simulation in VR. A 16-participant study, which presented variations of a collaborative block stacking task, showed that implicit techniques enable efficient task completion and task parallelization, and should be augmented with explicit mechanisms to provide users with fine-grained control.

**CCS CONCEPTS • Human-centered computing ~ Human computer interaction (HCI) ~ Interaction paradigms ~ Collaborative interaction**

## 1 INTRODUCTION

In human-robot collaboration (HRC), humans and robots work together to achieve a common goal. For example, in a simple industrial assembly task, a human-robot team builds a structure by picking up and placing objects in the environment. Collaboration in these tasks can be crucial especially when it leverages the unique strengths of the human (e.g., dexterous manipulation) and robot (e.g., precise placement of objects). We envision human-robot collaboration where the human and robot seamlessly work together as equal partners. One of the many challenges in human-robot collaboration is determining what actions the robot should take when collaborating with a human teammate, while avoiding conflicts such as picking up the same object – a problem known as task allocation [14]. Contemporary approaches to task allocation can include pre-planning the robot's [20] and potentially, the human's actions [26]. While pre-planning can reduce human cognitive load during task execution, it may come at the expense of human agency [12]. Humans may want flexibility in task completion due to many reasons – varying preferences, factors affecting their performance such as inherent ability or fatigue, factors relating to the robot's performance such as its inability to pick up certain objects, and unforeseen environmental factors. Approaches for more flexible task allocation involve online planning, where the robot adapts to the user's actions and selects actions that complement them [37]. However, online planning typically requires an initial shared plan to be finalized before task execution which may not always be possible.

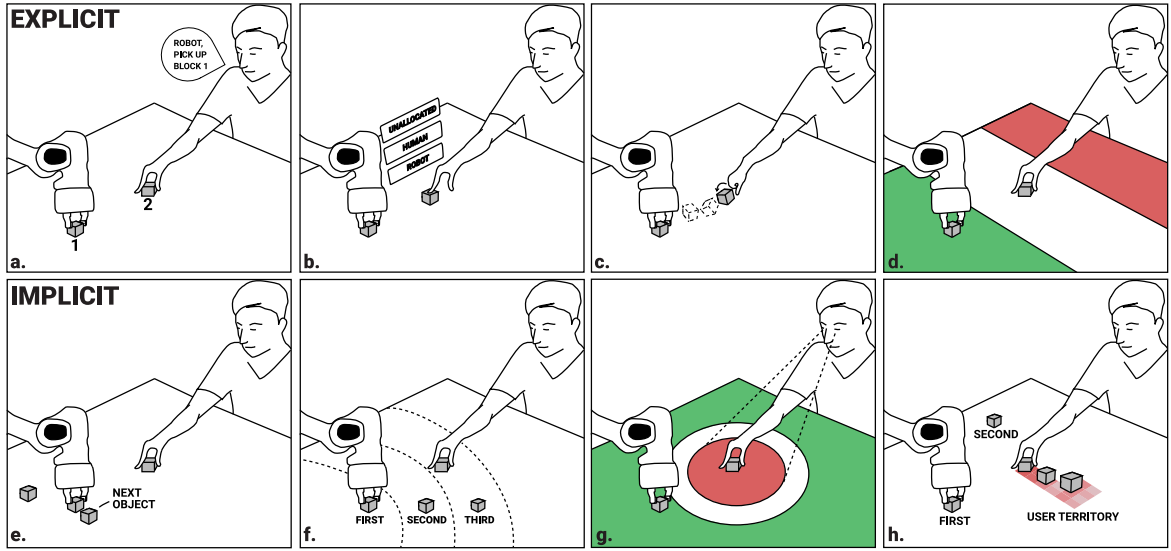

Figure 1: We envision task allocation techniques for ad-hoc collaboration that exploit coordination behaviors from human collaboration. Illustrated is our proposed design space, with explicit techniques where allocation occurs through user input and implicit techniques where the robot makes allocations through a heuristic. Explicit techniques include a. Voice where objects are allocated verbally, b. Menu which leverages spatial interaction, c. Subtle Relocation where objects are assigned if they are pushed towards the robot, and d. Fixed Territories where objects are allocated by moving them between territories. Implicit techniques include e. Proactive where objects nearest to the robot's gripper are selected first, f. Distance where objects are assigned according to their distance from the robot's base, g. Gaze where the user's eye gaze creates territories that the robot avoids picking objects from, and h. Proximity where the user's activity in a region creates territories that the robot avoids picking up objects from.

There are also many tasks in which pre-planning is not possible – creative tasks like completing jigsaw puzzles and cooking – where collaboration is *ad-hoc*, and task allocation is pursued through simple coordination behaviors. For example, akin to *"Put-that-there"* for graphical interfaces [3] the user could explicitly request the robot's assistance verbally, [2] the robot could monitor the user and *react* by intervening when assistance is needed [29], or the robot could take *proactive* actions instead of reacting to the user or requiring explicit user input [18]. This paper focuses on expanding the range of coordination behaviors to facilitate task allocation in ad-hoc human-robot collaboration with a specific focus on tabletop settings. We take inspiration from studies of human collaboration where team members incorporate coordination behaviors such as exhibiting an intuitive sense of what others are doing, understanding how the workspace is shared amongst team members [23], and utilizing subtle and spatial cues [1].

Through an understanding of coordination behaviors demonstrated in human collaboration, we propose a design space of novel *task allocation* techniques – ways of specifying what the robot and user should work on in real-time. Some techniques are *explicit*, requiring the user to allocate objects through deliberate action while others are *implicit*, enabling the robot to make inferences based on a heuristic. For instance, an explicit allocation in human collaboration can be initiated by one team member telling another, *"Pick up the yellow block close to you."* An implicit allocation could occur when a team member picks up an object nearby and places it across the table where their teammate is working. We propose a design space with three new explicit techniques: *menu, subtle relocation*, and *fixed territories* and three new implicit techniques: *distance*, *gaze*, and *proximity*.

We envision these allocation techniques, which augment visualizations elements and spatial interaction, being applied to HRC mediated by digital tabletops and mixed reality. The proposed techniques were prototyped in a tabletop HRC simulation in VR and evaluated in a controlled user study with 16 remote participants who completed various configurations of a collaborative stacking task. The findings suggest that implicit techniques inspired by human collaboration can improve task performance and should be augmented with explicit mechanisms to provide users with finer control over task allocation.

This work is an important step towards the design of intuitive task allocation which can reduce human burden in ad-hoc HRC, bringing us closer to the vision of robots and humans being equal partners. In summary, we make the following contributions: 1. a design space for task allocation techniques in tabletop HRC inspired by human collaboration, 2. the implementation of these techniques within a novel VR HRC simulation, and 3. a systematic evaluation of the techniques in a user study with 16 participants.

## 2 RELATED WORK

This section recaps prior approaches to task allocation – planning (offline and online) and coordination behaviors. Following this we summarize instances where HRI researchers have successfully incorporated coordination behaviors into HRC, as well as promising coordination behaviors that have not received much attention in HRC (territoriality and proxemics). Lastly, we review intuitive interfaces for robot control and end-user programming which motivated the design of our task allocation techniques.

### 2.1 Task Allocation

HRI researchers have proposed many approaches for task allocation. These include planners that rely on *a priori* task allocation performed offline, and online planners. Offline planners have been proposed which can plan at the team level, task level, and agent level while considering the skills needed by the agent (or robot) to perform them [20]. Other planners assign actions for the human and robot, by considering their unique capabilities, assigning actions to a specific member when it benefits from their touch [26]. However, these approaches [26,40] do not account for human preference and may reduce humans' sense of agency. On the contrary, researchers have demonstrated that humans prefer task allocation to be performed by autonomous planners as it is less burdensome [12]. In contrast to offline planners, online planners may not require a fixed action sequence be defined a priori. For instance there are planners which can adapt during the collaboration by choosing actions for the robot to complement their human teammate [37], although they may require an initial shared plan to be negotiated before task execution. Turn taking is another approach taken by task planners through which agents negotiate allocation during task execution [27].

Alternative approaches forego planning and propose coordination behaviors that human collaborators frequently use, to facilitate *ad-hoc collaboration*. These behaviors may include negotiating which objects to work on and actions to take by exchanging implicit or explicit cues, as well as observing the progress of teammates to resolve hesitation or frustration. Baraglia et al. propose three mechanisms [2] where: a human can request a robot's assistance explicitly, a robot can observe their human partner and react when they struggle, or a robot can proactively take actions to advance task completion without being prompted. Schulz et al. proposed the aforementioned as well as two further mechanisms: autonomous and human-commanded [31], and evaluated them in various HRC tasks. From these evaluations we surmise that users prefer to work with proactive robots but sometimes wish to control the robot more carefully especially when working on tasks that require *joint* or combined actions. We are encouraged by the success of coordination behaviors for task allocation in ad-hoc collaborative tasks and contribute a design space of task allocation techniques which employ coordination behaviors from human collaboration.

**2.2 Coordination Behaviors Inspired by Human Collaboration**

Researchers have observed how human collaborators coordinate their actions and appropriated them into the design of coordination for human-robot collaboration. For instance, Shah and Breazeal propose the Chaski task planner which adaptively plans robot actions based on what its human teammate is doing in a collaborative task. In addition, it incorporates several implicit and explicit coordination behaviors from human collaboration together such as providing status updates and acting to minimize the idle time of team members [36,39]. Robot object handover has been enhanced by incorporating the coordination strategies of waiting and slowing down to observe team members before handing them the next object [17]. Based on human studies of the use of communicative gestures, researchers have imbued robots with the ability to gesture to signify operations on objects for an assembly task [11], and to *reference* objects in different task contexts [30] (e.g. when blocks are far from the referrer or when there is no visibility). Hoffman and Breazeal found that a robot employing an anticipatory action selection mechanism by observing and predicting the human teammate's next action outperformed a naïve reactive action selection mechanism in a simulated HRC construction task [16]. Human gaze is another useful coordination behavior used by robots to resolve hesitation when humans are choosing between two similar-looking objects [29], and to proactively plan to pick up an object that a human is looking at but has not yet requested the robot to pick up [18].

Our interest is in adapting coordination behaviors from human collaboration as it relates to task allocation. We specifically focus on the practices of territoriality and proxemics. *Territoriality* is an implicit mechanism that explains how collaborators distribute resources for task completion on the workspace into different territories: *personal, group,* and *storage* [33]. *Proxemics* deals with how humans navigate social spaces with other humans [5,13,38]. The phenomenon has also been demonstrated when humans are in close proximity to robots [23] and engineered to enable socially acceptable robot navigation [22]. *Proxemic interaction* [1] enhances the notion of proxemics and wrestles with the idea that people's physical and spatial relationships with objects indicates and shapes the ways in which they use them. HCI researchers have exploited proxemic interaction to design media players in shared living spaces [1], public interactive whiteboards [21], and interactive public ambient displays [41]. However, to our knowledge, neither territoriality nor proxemic interaction have been incorporated in human-robot collaboration, particularly for task allocation. Territoriality and proxemic interaction form the basis for five of our proposed task allocation techniques.

**2.3 Intuitive Interfaces for Robot Control**

Task allocation in ad-hoc collaboration involves instructing the robot what to do at a given moment. This can be considered a problem of robot control, for which many intuitive interfaces exist. Sketch-based interfaces enable users to draw navigation paths or invoke immediate commands such as pausing or stopping [28]. User pointing is an intuitive mechanism to select objects for a robot to work on [24,25] and has been mimicked with technical solutions such as laser pointers [19]. Augmented Reality (AR) interfaces allow users to select workspace objects for robot manipulation through a tablet [9] and to teleoperate robots via an HMD [42].

HRI researchers have also designed tools for programming a robot to achieve a sequence of steps in an intuitive manner. GhostAR [6] allows users to record an interaction sequence where the robot and the user's actions are pre-specified. Later, the system can synchronize a live user's actions with pre-recorded user and robot actions. PATI utilizes a tabletop projection system to assist users in programming a robot to perform pick-and-place tasks [10]. Situated tangible robot programming lets users place physical blocks in parts of the workspace to instruct the robot to pick up and drop an object [34]. We propose techniques inspired by these interfaces to enable users to define real-time allocation sequences in HRC tasks.

## 3 DESIGN SPACE OF TASK ALLOCATION TECHNIQUES FOR TABLETOP HRC

Guided by our review of prior literature, we propose a design space of techniques for task allocation in tabletop HRC. Our design space consists of two main dimensions (technique class and allocation perspective) and one sub-dimension (workspace type). The dimensions as they map to each technique can be seen in Table 1.

***Technique Class:*** Techniques can be *explicit* – where an allocation is made through a deliberate action taken by the user or *implicit* – where the robot proactively allocates objects to itself based on a heuristic.

***Allocation Perspective:*** *Object-centric* allocations occur by direct interaction with an object. *Robot-centric* allocations are made by considering the relationship between objects and the robot. *Workspace-centric* allocations consider how objects are located with respect to the workspace they occupy.

***Workspace Type:*** For techniques that utilize a *workspace-centric* allocation perspective, the workspace can be either *static*, remaining constant over time, or be *dynamic*, changing over time depending on the context.

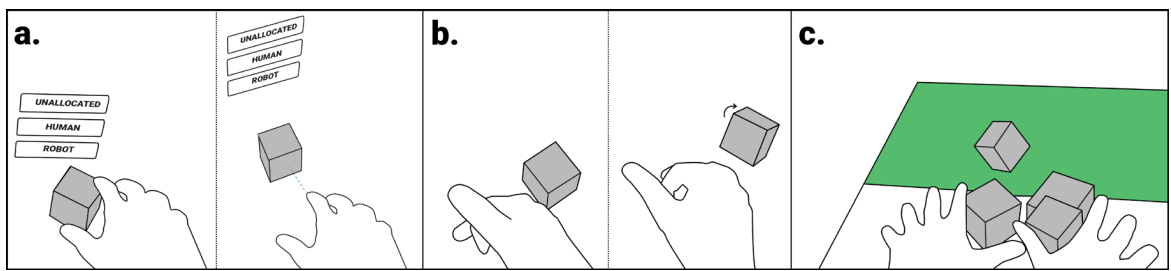

Figure 2: Explicit allocation techniques: a. Menu – touching and pointing to allocate objects; b. Subtle Relocation – pushing an object towards the robot to allocate it; c. Fixed Territories – moving objects into the robot's territory to allocate them.

### 3.1 Explicit Techniques

Explicit techniques enable task allocation through deliberate user action. The robot only attempts to pick up an object after it has been allocated. We prototyped four explicit allocation techniques, visualized in Figure 2:

***Voice (Baseline):*** With this technique, the user can assign an object to the robot by reciting the number that appears atop the object. The technique is a re-implementation of an existing technique that prior work has proposed to request assistance from a robot [2] and to command it [18], serving as a baseline for comparisons.

***Menu:*** Using this technique, the user allocates objects through a familiar menu metaphor that is ubiquitous. Allocations are made by touching and holding an object or by pointing at an object via the index finger until a menu appears, which allows them to allocate an object to themselves, the robot, or cancel a previous allocation.

***Subtle Relocation:*** This technique allows the user to assign an object to the robot by pushing it in its direction through direct manipulation. Likewise, the user can assign an object to themselves by pulling it towards their direction. The technique requires a successful gesture to be detected and for the object to move in the direction of the user or robot.

At a conceptual level, the technique exploits the notion of proxemic interaction - objects on the table can recognize their movement with respect to the user's and robot's locations to determine who should work on them. It leverages the *movement* dimension of proxemic relationships [1] whereby a change in an object's position (velocity) determines whether an object should be assigned to the robot or whether the user is simply moving objects around the table.

*Fixed Territories:* This is a technique inspired by tabletop territoriality [33], an implicit mechanism through which collaborators naturally partition the workspace into different regions, some belonging to each user (personal), others belonging to the group, and additional regions where task-related objects are stored. We implemented personal and group territories in fixed locations of the table depending on where the user and robot are situated at the start of the task. The territories are *robot* - regions closest to the robot, *human* - regions closest to the user, and *group* - regions between the user and robot. Objects can be allocated to the robot or the user by manipulating them into the respective territory and can be unallocated by moving them into the group's territory. The territories are visualized on the tabletop with green regions indicating the robot's territory, red regions denoting the user's territory, and white regions signifying the group's territory.

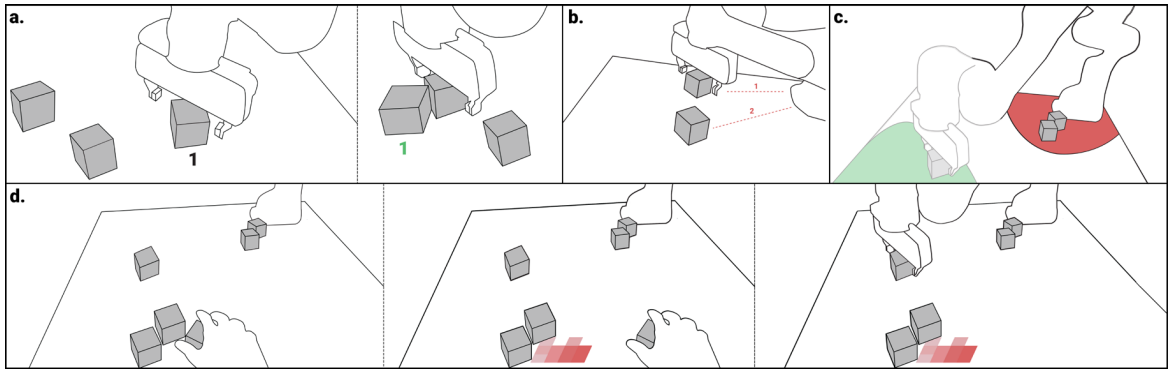

Figure 3: Implicit allocation techniques: a. Proactive – robot picks up the object labelled "1" and places it at the target. Then, it picks up the object closest to its gripper; b. Distance – robot picks up the object closest to its base, starting with the object labelled "1" and then "2"; c. Gaze - user looks at the robot's base, causing it to become red (user's territory) and the robot picks up object on the opposite side where the user isn't looking (robot's territory); d. Proximity - user picks up an object and moves it to a new location, causing its initial position to become the user's territory. Then, the robot avoids picking up any objects near the user's territory and instead picks up an object from the group's territory.

### 3.2 Implicit Techniques

This class of techniques assists the robot in making its own task allocations by continuously evaluating a heuristic to select an object to work on. We propose four implicit allocation techniques, seen in Figure 3:

*Proactive (Baseline):* This technique leverages the *position* dimension of proxemic relationships in *relative* terms [1]. It is a re-implementation of the proactive robot mechanism described by Baraglia et al. [2] where the robot assists the user by picking up objects closest to where it has recently completed an activity (such as placing an object). As the gripper is empty, the robot can find a nearby object to assist the user. This technique serves as a baseline to compare the proposed implicit techniques.

*Distance:* This is a technique that borrows the notion of proxemic relationships between objects; here we consider the relationship between the robot's base and objects in the workspace. The *position* of an object in *absolute* terms [1] determines whether an object should be picked up by the robot – the robot allocates objects to itself depending on their distance from its base. Unassigned objects that are closest to the robot's base are prioritized while objects furthest from the robot are allocated later. The user can influence what the robot picks up by changing the objects' relative positions from its base.

***Gaze:*** This technique employs the proxemic interaction concept of *attending to other surrounding objects and devices* [1] through which the user is not attempting to directly control the system (robot in this case); rather they direct their attention to objects in the workspace. User gaze in a region facilitates the creation of dynamic user and robot territories [33]. The robot assigns objects to itself based on whether they are in its territory (the priority) or in the group's territory, while avoiding picking up any objects that are in the user's territory. We used head tracking from the VR headset to calculate a "gaze" score - a measure of the extent to which the user is looking at an area of the workspace. To determine which area, the table is discretized into a rectangular grid, where each tile corresponds to a region in the workspace. The gaze score of each region is *averaged* over a fixed timeframe (a system parameter) which is used when the robot is allocating objects to itself. When the robot makes an allocation, it calculates an *average* of the gaze scores (which are already averages) of regions nearest to every object. Then, the robot selects the object with the lowest gaze average signifying that the user is not working on that object or objects near it. A heatmap visualization helps the user understand how their gaze affects the formation of territories, by displaying red regions to indicate the user's territory, green regions to indicate the robot's territory, and white regions which belong to the group.

***Proximity:*** This technique is also based on tabletop territoriality [33]. The robot allocates objects in regions close to or within its territory first, followed by objects in the group's territory, and avoids objects close to or within the user's territory. At the start of the task, the entire workspace belongs to the group. Each time the human or robot picks up an object from an area of the workspace (its initial location) and places it in another, the regions nearest to the object's initial location are infused with a *proximity score* to signify that these regions are being used by the user or robot respectively, leading to the formation of territories. When there is no further activity in a region, these scores decay back to becoming the group's territory. Further, a territory belonging to the robot can be overridden if the user picks up an object from that region (infusing it with a user proximity score). When the robot makes an allocation, it calculates the average proximity score of regions nearest to each object and picks the object with the highest robot proximity score if one exists, or an object with a low proximity score (in group territories) and avoids objects with a high user proximity score above a certain threshold (a system parameter). These territories are also visualized via a tabletop heatmap.

Table 1: Mapping the proposed techniques to the dimensions of the allocation techniques design space.

| Technique | Voice | Menu | Subtle | Fixed | Proactive | Distance | Gaze | Proximity |
|---|---|---|---|---|---|---|---|---|
| *Class* | Explicit | Explicit | Explicit | Explicit | Implicit | Implicit | Implicit | Implicit |
| *Perspective* | Object | Object | Robot | Workspace | Robot | Robot | Workspace | Workspace |
| *Workspace* | x | x | x | Static | x | x | Dynamic | Dynamic |

## 4 EVALUATION TESTBED

To evaluate our techniques, we prototyped a human-robot collaboration simulation in VR. Our rationale for using VR was partially influenced by the COVID19 pandemic as testing in shared physical spaces was a challenge. However, a VR implementation allowed us to consider the design of techniques that exploit the benefits of digital tabletops and mixed reality, such as the ability to visualize what the robot is doing or providing a clear mental model of certain techniques with heatmaps.

Our simulation integrates the Unity game engine[1] and ROS[2] facilitated by ROS#[3] through WebSocket. The Unity environment consists of a tabletop workspace that is shared by a user and a collaborative robotic arm. The user interacts with the simulation environment through a VR controller or the Leap motion. The robot's physics are simulated by the ROS simulation engine, Gazebo, through a URDF model provided by the robot's developers (Franka Emika). This robot URDF is also imported to Unity and serves as a visualization for the user, with its position and orientation mirroring its Gazebo equivalent through a ROS topic. Unity also provides constant updates about the positions and orientations of all tabletop objects to Gazebo through a ROS topic which is utilized for motion planning. When an object is allocated to the robot and needs to be picked up, Unity requests a motion plan from the MoveIt framework through a ROS service. Upon generating a motion plan, MoveIt begins to play the trajectory and the robot's position and orientation are mirrored in Unity. The system is designed so that the server and client can run on machines in different networks which helped us to conduct the remote user studies.

The allocation techniques were implemented in Unity and leveraged a task allocation manager to request the robot to pick up and place objects. The task allocation manager interacted with MoveIt to request motion plans. All techniques worked autonomously, but we employed Wizard-of-Oz [8] for the *voice* technique in the user studies to prevent inconsistent performance from differing input and output audio participant hardware. With the *voice* technique, when the user verbally requests the robot pick up an object, the wizard presses the corresponding number on their keyboard causing the robot to make a motion plan towards that object. Please see supplementary materials for implementation details of all techniques.

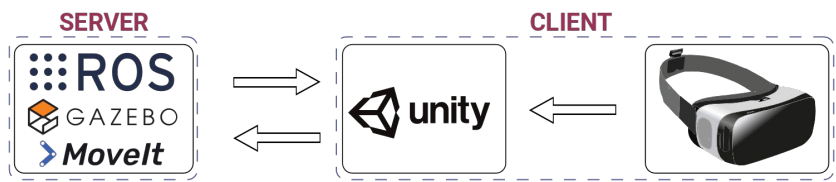

Figure 4: Evaluation testbed where: the Client executes a Unity-based simulation which the user interacts with through a VR headset; and the Server which receives information about objects on the workspace as well as user commands through ROS# which are processed by the Gazebo simulation engine and sent to MoveIt to prepare motion plans. The server and client can run on the same or different machines on different networks.

## 5 EVALUATION

We designed a controlled lab experiment to assess the utility and usability of our newly proposed allocation techniques in a single robot-single user setting. Although the explicit techniques provide a clear model of interaction for task allocation, they require manual input and can be cumbersome for users. On the other hand, users may have a more difficult time building mental models of the implicit techniques but upon achieving this could be freed up to work as they naturally would with human teammates. Ideally, the user would be able to complete their task without expending much effort and time tending to the robot. On the other hand, a poor technique would force the user to waste considerable time directing the robot (in the explicit case) or waste time waiting for the robot to make the right allocations (in the implicit case). Though some of the techniques in our design space have been tested in the past (the baseline techniques - *voice* and *proactive*), there has never been a comprehensive evaluation of the relative benefits and drawbacks of the classes of techniques they represent, namely explicit and implicit techniques.

---

[1] https://unity.com/
[2] https://www.ros.org/
[3] https://github.com/siemens/ros-sharp

To assess the techniques' performance, we chose a block stacking task, which is commonly employed in HRC studies. The task was modelled to resemble an equal partnership where both team members had an equal number of blocks that only they could place at the goal locations. Our interest was in studying how the techniques would perform when the task necessitates varying levels of human-robot coordination, so we designed two tasks - a *coupled task* where the stacking structures required the placement of the user's and robot's blocks in alternating fashion, and a *decoupled task* where the stacking structures for the user and robot were separate and could be completed independently. Since our techniques differ in allocation perspective with some being *object-centric*, *robot-centric*, and *workspace-centric* we were also curious as to whether the placement of the user's and robot's objects would affect technique performance; so, we manipulated whether user's and robot's objects on the table were *scattered* or *sorted*.

### 5.1 Task Rules

In the stacking tasks, only the user could place yellow blocks at the goal locations and only the robot could place black blocks at the goal locations, which users were informed of prior to the study. However, the robot was not provided this knowledge *a priori* meaning that it only realized that a yellow block could not be picked up upon attempting it. This constraint was designed to model tasks where the robot is not aware of how its capabilities map to the task at hand, and the user has no mental model of the robot's capabilities - tasks such as those involving the manipulation of oddly shaped objects or even soft objects. When using implicit techniques, the robot could attempt to pick up each yellow block once if it were allocated, but it would never re-attempt to pick up a yellow block upon failing. Hence it was the user's responsibility to ensure that the robot did not waste time attempting to pick up yellow blocks. Upon allocation, in our implementation, the robot could pick up and place all black blocks with a success rate similar that of the average user.

### 5.2 Participants

Seventeen participants aged 18 to 45 (4 females, 13 males) were recruited via research networks and forum posts (e.g., Reddit). Participants were primarily HCI researchers (11 out of 17) and included some VR enthusiasts (6 out of 17). We dropped one female participant from the analysis due to technical difficulties when conducting the study. All participants had VR experience while 50% had used VR for over three years, and used different VR HMD (HTC Vive: 2, HTC Vive Pro: 3, Oculus Rift: 5, Oculus Quest: 4, and Valve Index: 2).

### 5.3 Measures

Data from the simulation was collected when participants completed each trial. These included: 1. task completion time, 2. user trial time, segmented into the time the user spent idling, manipulating objects towards the goal, maneuvering objects without placing them at the goal, and allocating objects when using an explicit technique; 3. robot trial time, segmented into the time the robot spent idling, manipulating objects towards the goal, and reaching objects; 4. number of touches by the user on an object, segmented by touches for allocation, manipulating objects, and maneuvering objects; 5. number of attempts by the robot to pick up a yellow object (errors); and 6. a log of the user's and robot's actions at each time step.

Participants also provided responses to a questionnaire which asked for feedback about each allocation technique. The questionnaire gathered feedback on: 1. the ease of the technique, 2. the efficiency of the technique, 3. the level of control over which objects the robot attempted to pick up, and 4. the level of parallelization over task completion. They also provided feedback on team fluency [15]: 1. whether the human-robot team worked fluently together, 2. whether the human-robot team's fluency improved over time, and 3. whether the robot contributed to the fluency of the interaction. Lastly participants provided some open-ended comments about each technique and how they might modify it and ranked the techniques in order of preference.

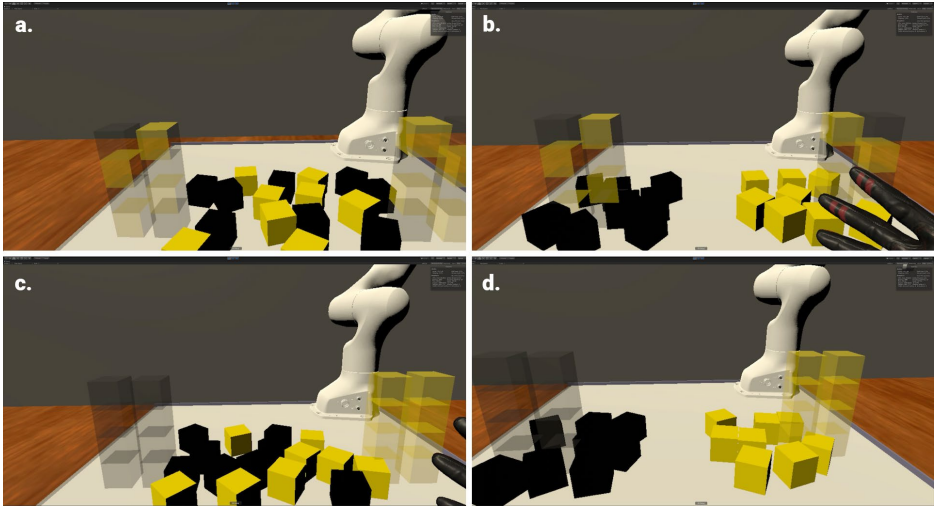

Figure 5: The 2 x 2 experimental block participants completed when trying each interaction technique, in which the stacking task can be coupled (a, b) or decoupled (c, d), and objects are either scattered (a, c) or sorted (b, d). In coupled tasks, structures require the placement of both the user's and robot's blocks while decoupled tasks can be completed independently.

### 5.4 Procedure

The study was conducted remotely due to the COVID19 pandemic. Participants completed the study at home using their own hardware while we observed them through a video call. They deployed the Unity simulation on their machines which communicated with the ROS programs running on the experimenter's machine. Prior to the study, participants completed a consent form and a demographic survey. We first introduced them to the experiment, after which they became accustomed to the controls by completing a pre-study trial where they stacked two structures of four yellow blocks each without the robot's help. Then, participants watched a video showcasing the current technique and completed a practice trial where they again stacked two structures of four yellow blocks. Next, they completed the 2 x 2 experimental block (seen in Figure 5), where they encountered every combination of type of stacking task and object placement type in a randomized fashion. Regardless of the combination, in each task the user and robot completed four structures each consisting of four blocks, of which the user completed half (yellow blocks) and the robot completed the other half (black blocks). At the end of each experimental block, participants completed the short post-condition questionnaire. Participants ranked the techniques after the study.

### 5.5 Study Design

Our study employed a within-subjects experimental design and manipulated 3 variables: 1. allocation techniques (the eight described), 2. type of stacking task (coupled versus decoupled), and 3. object placement (scattered versus sorted).

An 8 x 8 balanced Latin square was used to expose participants to different orderings of the techniques. Within each technique condition, participants were exposed to every combination of the 2 x 2 = 4 combinations of stacking task and object placement once in randomized order.

## 6 FINDINGS

We received data from 32 trials each from 16 participants (excluding practice trials). Six trials were not properly logged due to data corruption, leaving 506 trials (16 x 32 – 6 = 506). No outliers were rejected. The findings emphasize comparisons between the allocation techniques in the presence of the other variables - task type and object placement, for the following data: 1. performance measures such as task completion time and robot errors, and 2. objective measures of team fluency including user idle time, robot idle time, and concurrent activity percentage. We also analyzed users' subjective perception of 1. technique performance and 2. team fluency. Repeated measures ANOVA tests were conducted to examine the quantitative data, with Bonferroni corrections for multiple comparisons and Greenhouse-Geisser corrections when the test for sphericity was not met. In the plots, grey backgrounds indicate explicit techniques, while error bars represent standard error. Horizontal bolded lines indicate pairwise significance within a technique between conditions, while letters atop the bars indicate pairwise significance between techniques.

### 6.1 Technique Performance

*Completion Time:* Overall, the findings suggest that implicit techniques performed nearly as well as the baseline *voice* technique, while the *fixed* explicit technique also demonstrated promise. However, the object placement type and task types influenced how the techniques impacted task completion.

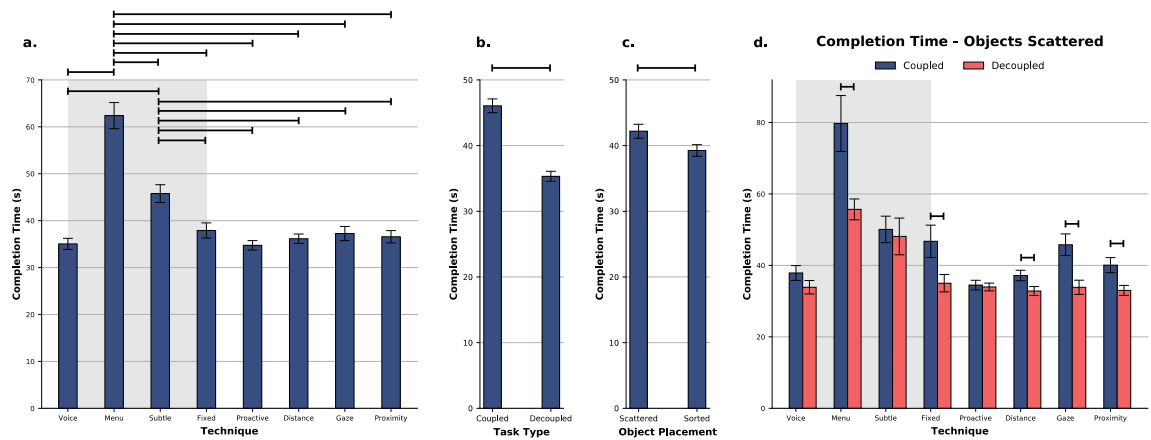

Figure 6: Plots showing - a. completion time versus allocation technique; b. completion time versus task type; c. completion time versus object placement type; and d. comparison of completion times for each allocation technique by task type for the case where objects were scattered. Explicit techniques are shown in grey for plots a. and d.

Figure 6a, b, and c show simple plots of completion time by technique, task type, and object placement. There was a significant effect of technique ($F(4.56, 282.710) = 33.664$, $p < .0005$), task type ($F(1, 251) = 77.613$, $p < .0005$), and object placement type ($F(1, 251) = 5.870$, $p = .016$) on completion time. Pairwise comparisons showed that: 1. the *menu* was slower than all techniques, and *subtle* was slower than *voice*, *fixed*, and all implicit techniques, 2. the *coupled* task took longer than the *decoupled* task, and 3. the task took longer when objects were *scattered* versus when they were *sorted*.

There was a significant interaction between technique, task type, and object placement type (F(2.777, 38.882) = 2.941, p = .049). Further inspection revealed an interaction between allocation technique and task type but only when objects were *scattered* (F(2.509, 35.121) = 3.140, p = .045) as Figure 6d shows. Pairwise comparison of the same technique between tasks (Figure 6d blue vs. orange) revealed significant increases in completion time for the *menu*, *fixed*, *distance*, *gaze*, and *proximity* techniques in the *coupled* task compared to the *decoupled* task, suggesting that participants took longer to complete the task when a higher level of human-robot coordination was needed, and objects were *scattered*.

**Robot Errors:** Except for the *subtle* technique, explicit techniques led to fewer robot errors while the implicit techniques led to a larger number of robot errors. The number of robot errors when using implicit techniques changed with the type of task and object placement as Figure 7 illustrates.

We observed a significant interaction between technique and task type (F(2.992, 41.892) = 4.070, p = .013) and a significant interaction between technique and object placement type (F(4.047, 56.659) = 4.051, p = .006). Pairwise comparison of the same technique between tasks (Figure 7b) revealed that the *distance*, *gaze*, and *proximity* techniques caused more robot errors in the *coupled* task compared to the *decoupled* task. One explanation might be that these techniques, which are robot- and workspace-centric required users to keep track of the structure they were building, while simultaneously scanning and moving yellow objects that the robot might target next as these techniques are all autonomous.

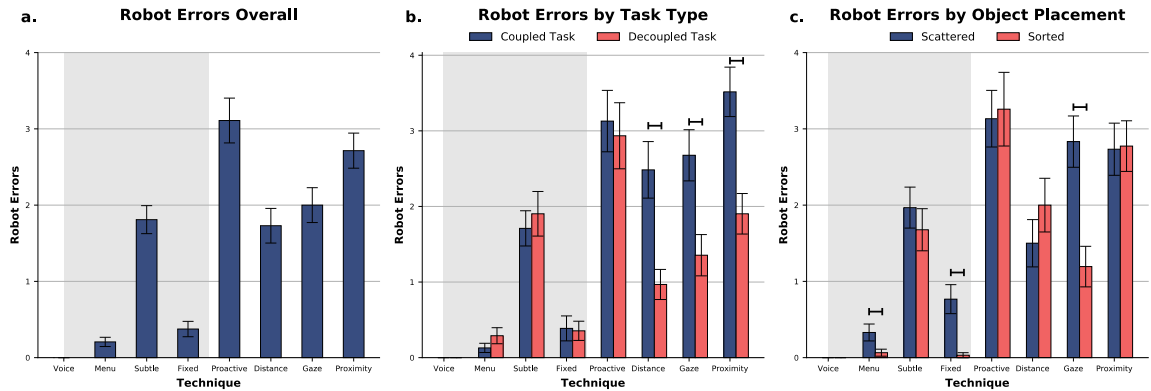

Figure 7: Plots showing - a. robot errors by allocation technique; b. robot errors by task type for each technique; and c. robot errors by object placement type for each technique. Explicit techniques are highlighted in grey.

Pairwise comparison of the same technique between object placement type (Figure 7c) showed that the *menu*, *fixed*, and *gaze* techniques caused more errors when the objects were *scattered* compared to when they were *sorted*. *Gaze* became much harder to use when objects were *scattered* possibly because looking at an area of the table meant that another area of the table containing both yellow and black objects was in the robot's territory, causing it to make more incorrect inferences about what to work on (the yellow blocks).

### 6.2 Objective Measures of Team Fluency

In HRC team fluency can be quantified by calculating the total percentage of time spent by the user and robot idling, and the amount of concurrent activity [15]. Figure 8a and b illustrate the total percentage of time spent by the human-robot team on different action types, including idling, taking goal-related actions, and taking overhead actions. Overhead time was calculated for the user by counting the time spent moving objects around the workspace without placing them at the goal, while overhead time for the robot included time spent reaching objects that it was not supposed to pick up (yellow blocks). HRC literature suggests that idle time percentage should be low for fluent collaboration while concurrent activity should be high.

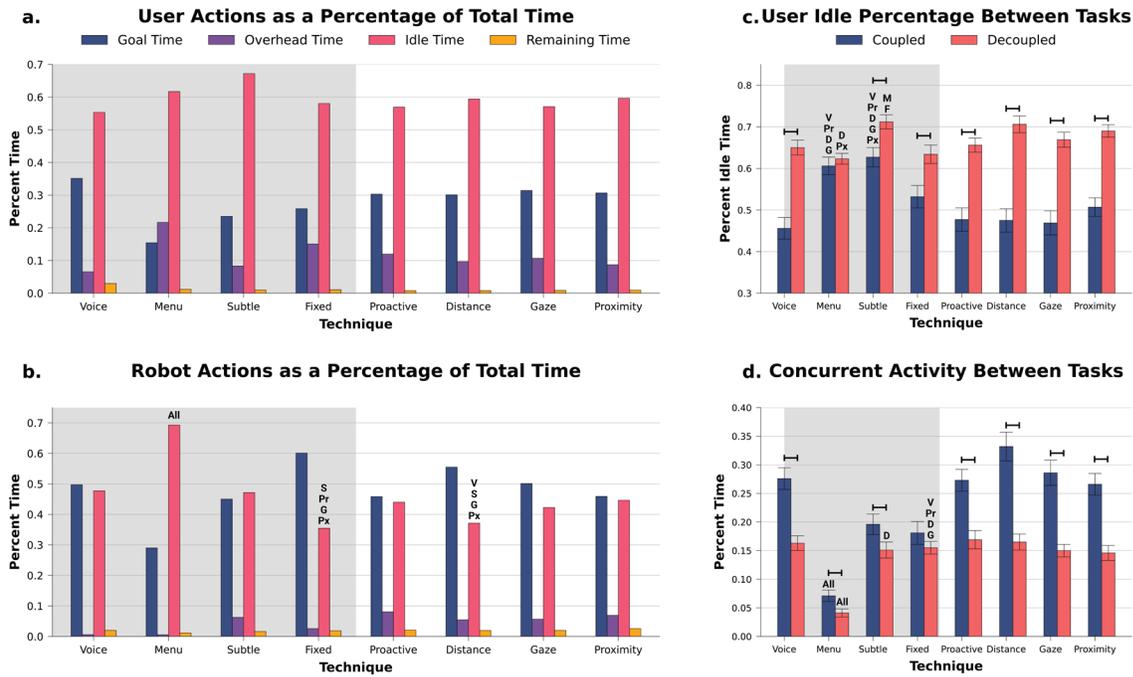

Figure 8: Plots showing percentage of time spent on goal-related actions, overhead actions, and idling by the - a. user and b. robot; c. user idle time percentage by task type for each allocation technique; d. concurrent activity percentage by task type for each allocation technique. Explicit techniques are highlighted in grey. Letters above bars indicate pairwise significance between techniques. Legend – V: Voice, M: Menu, S: Subtle, F: Fixed, Pr: Proactive, D: Distance, G: Gaze, Px: Proximity.

*User Idle Time Percentage:* Overall, idle time percentage was consistently high despite the technique (Figure 8a) but varied between task types. There is a significant interaction between technique and task type ($F(7, 98) = 6.487$, $p < .0001$), as seen in Figure 8c. Pairwise comparison of the same technique between tasks demonstrated that users idled for significantly longer in the *decoupled* task versus the *coupled* task with all techniques except the *menu* where idle time was always high. This is surprising as the *coupled* task required more coordination increasing the likelihood that the user would be idling while waiting for the robot, but this may be explained by users' inherent manipulation speed advantage over the robot, causing them to complete their structure in the *decoupled* task much faster, leading to more idling before the task was completed by the robot.

Pairwise comparison between different techniques for the *coupled* task showed that the *menu* led to more idling than the *voice*, *proactive*, *distance*, and *gaze* techniques, while the *subtle* technique led to more idling than the *voice* and all implicit techniques. Pairwise comparison of techniques for the *decoupled* task revealed that the *subtle* technique led to larger idle percentages than the *menu* and *fixed* techniques, and the *menu* resulted in lower idle percentages than the *distance* and *proximity* techniques. The menu always led to long idle times which did not drastically increase in the *decoupled* task unlike with the other techniques possibly because users had to spend considerable time triggering the menu and selecting an option to allocate each object. This is evident in Figure 8a where overhead time for the *menu* is over 20%.

***Robot Idle Time Percentage:*** There is a significant effect of technique ($F(7, 98) = 80.656, p < .0005$) and task type ($F(1, 14) = 9.519, p = .008$) on robot idle time percentage. Pairwise comparisons between techniques showed that: 1. the *menu* led to larger idle percentages compared to all techniques, 2 the *fixed* technique led to lower idle percentages compared to the *subtle*, *proactive*, *gaze*, and *proximity* techniques, and 3. the *distance* technique led to lower idle percentages than *voice*, *subtle*, *gaze*, and *proximity* techniques. Interestingly the *fixed* technique led to lower idle times for the robot possibly because users could allocate multiple objects quickly, so the robot always had an available action. Similarly, the heuristic for the *distance* technique always found an object if the task was not complete, keeping the robot occupied. Pairwise comparison revealed that the *coupled* task led to longer idle time percentages than the *decoupled* task. This contrasts the result from user idle time percentage where users idled more in the *decoupled* task, suggesting they may have taken additional actions to facilitate robot task completion in the coupled task rather than idled. In addition, the total idle percentages for the robot are lower than the user, perhaps because the robot took longer to execute all its actions, from reaching an object to placing it at a goal location, when compared to the user. Another explanation may be that users had rich interactions with the blocks, such as being able to use two hands or to push blocks around, whereas the robot could only maneuver one block at a time, limiting its efficiency.

***Concurrent Activity Percentage:*** As Figure 8d shows, there is a significant interaction between technique and task type ($F(7, 98) = 4.771, p < .0005$). Pairwise comparison of the same technique between tasks showed that concurrent activity was higher for the *coupled* task than the *decoupled* task except in the case of the *fixed* technique. Despite requiring more coordination (which increases the likelihood of idling), it is interesting that concurrent activity was higher in the *coupled* task, which might once more be explained by the robot taking longer to complete each action than the user, leading to less opportunities for concurrent actions in the *decoupled* task.

Pairwise comparison between techniques for the *coupled* task revealed that concurrent activity was lower for the *menu* compared to all techniques. Pairwise comparison between techniques for the *decoupled* task also showed that the *menu* resulted in lower concurrent activity than all techniques, the *subtle* technique led to lower concurrent activity than the *distance technique*, and the *fixed* technique led to lower concurrent activity than the *voice*, *proactive*, *distance*, and *gaze* techniques. This hints that explicit techniques might have led to lower concurrent activity, especially for the coupled task, due to the overhead required for each allocation, which was less of an issue for the *voice* technique due to its multimodal nature.

## 6.3 Subjective Preferences on Technique Performance and Team Fluency

Figure 9 shows participants' responses to the questionnaire. Their responses to each explicit (4) and implicit (4) technique were grouped, yielding 64 responses for each technique category (16 participants x 4 explicit/implicit techniques). To contrast the differences between the explicit and implicit techniques in participants' responses, we conducted Wilcoxon signed-rank tests on these pairs of 64 responses for every question about *technique performance* and *team fluency*.

***Technique Performance:*** There was a significant effect of technique type on user perception of *being in control* ($z = -5.264, p < .0005$), with explicit techniques favored in 47 of 64 responses (Median: 6) versus 9 of 64 responses favoring implicit techniques (Median: 4). We found a significant effect of technique type on user perception of *being able to parallelize* task execution ($z = 1.964, p = .05$); 32 of 64 responses favored implicit techniques (Median: 6) while 20 responses favored explicit techniques (Median: 5). There were no significant differences in participation perception of ease and efficiency by technique type.

*Team Fluency:* We did not observe significant differences in participant perception of robot fluency, fluency over time, or robot contribution by technique type.

*Technique Preference:* The implicit techniques were generally liked but did not receive more than 1 vote for the most preferred in the case of the proactive, distance, and gaze techniques. For instance, the proximity technique received 5 votes for the second most preferred technique. In contrast, perceptions of the explicit techniques varied more. Amongst all techniques, 11 out of 16 participants most preferred the voice technique, and 6 out of 16 participants most preferred the fixed technique. On the other hand, the least preferred technique was menu, as stated by 13 out of 16 participants.

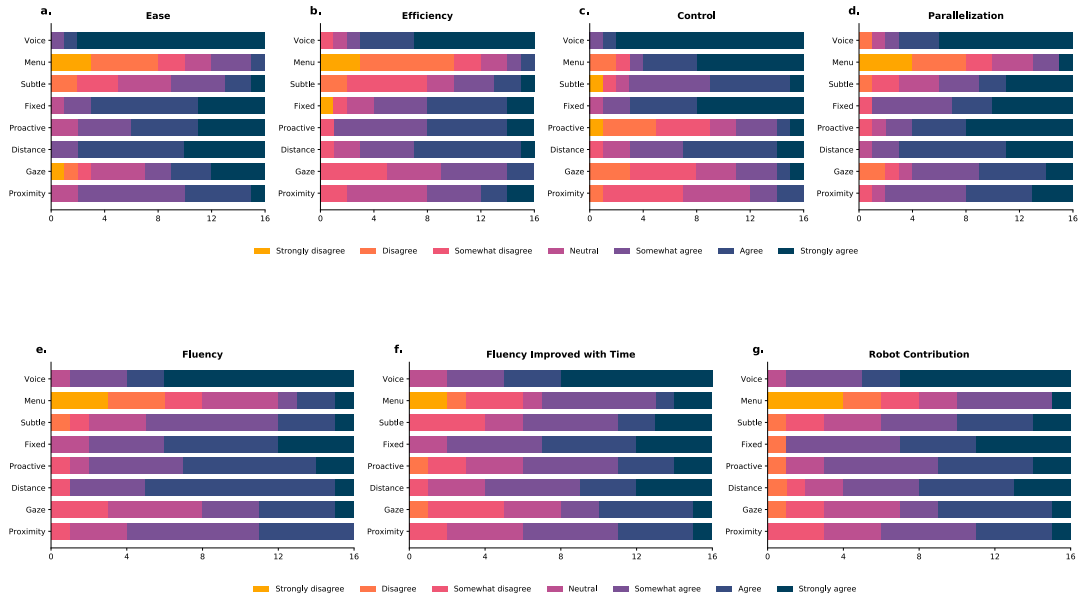

Figure 9: Plots showing the number of 7-point Likert scale responses to questions on technique performance - a. ease of use, b. efficiency, c. control, and d. parallelization; and fluency - e. team fluency, f. changes in team fluency over time, g. robot contribution.

## 7 DISCUSSION

### 7.1 Implicit Versus Explicit Techniques

The results suggest that task completion was aided by implicit techniques since they enable proactive task allocation without the user having to expend significant effort and time. With explicit techniques, users spent more time allocating objects to the robot, which contributed to increased overhead time, reduced concurrent activity, and in turn increased completion time, contributing to less fluent HRC. Though the *voice* technique performed well in this regard due to its multimodality, we could not quantify overhead time as we used Wizard-of-Oz. Further, there is additional overhead for the explicit techniques which currently appears as idle time – time spent by users waiting to see whether their allocation action was successful. This was especially evident when using the *subtle* technique, where gauging the pushing threshold was difficult and made the gesture harder to execute, particularly in the presence of yellow and black objects. The negative effects of the explicit techniques were more evident when working on the *decoupled* task where independent action was desirable but not possible due to the user having to constantly allocate objects for the robot. Despite the weakness of several explicit techniques, the *fixed* technique performed well and was well liked by participants as it offered a simple model to allocate objects and keep track of the robot's activities.

Although the implicit techniques performed well, they also incur overhead, some which resulted from participants treating the implicit technique explicitly while building a mental model of it. For example, when using the *distance* technique P4 remarked, *"At the start I was pushing blocks near it but realized didn't have need to do that as it would find the blocks anyway."* Other times participants felt that they had to maneuver objects for the robot to efficiently do its job, detracting from completing their own task. P12 said, *"My tactic was to look at a fixed point and bring all blocks of mine towards it and moved others away from it. This felt more efficient and I wasn't giving anything to the robot aside from the initial stage."* Some participants moved their objects to a region they were looking at to prevent the robot from accidentally grabbing them. P5 said, *"I again spent some time preparing the work areas by bringing my blocks over to my side, while making sure that the black blocks were not in red areas."* Additional overhead resulted from the robot mistakenly allocating yellow blocks, which happened often with the implicit techniques. Though it was not hugely detrimental to this task, incorrect robot inferences may undermine task execution if they are not addressed.

There exists a clear tension between allocation techniques offering control over task allocation and providing the ability to parallelize. Explicit techniques offer a strong sense of control over task allocation, restoring agency to the user while implicit techniques facilitate parallelization and concurrent activity. However, we are encouraged to see that some techniques such as *voice*, *fixed*, and *distance* seem to achieve both to varying extents - *fixed* allowed users to quickly allocate objects freeing them to focus on their task, while *distance* enabled proactive robot behavior with little user allocation effort.

### 7.2  Designing Intuitive Allocation Techniques for HRC

Our results suggest that allocation techniques inspired by simple coordination behaviors such as territoriality and proxemics have straightforward potential. Several of our techniques achieved similar results to *voice* despite its performance representing a best-case scenario (the Wizard's voice recognition was perfect, and the robot's task consisted of only 8 blocks which were easy to keep track of). Though implicit techniques were generally well received as evidenced by participants' subjective preferences (Figure 9), not all of them were universally liked. *Gaze* and *proximity* were found to be overbearing at times. P4 felt that when yellow blocks were clumped together, picking up an object did not prevent the robot from picking up yellow blocks in periphery regions to the newly spawned user territory. In this case, the area of effect was too small. In contrast, too large an area of effect when black and yellow blocks were clumped together would have prevented the robot from picking up any black blocks if the user picked up even one yellow block from the region. The *gaze* technique also suffered from having a large area of effect.

Sometimes the user's gaze would modify the table's territories such that the robot could not take further actions, which was especially apparent when yellow and black blocks were scattered. This often forced users to look away, which did not feel intuitive *("I had to look at the blocks to get them but looking at them triggered an area there that robot should not pick up blocks from." [P11])*. Even when the user and robot worked on separate structures in the decoupled task, the user was unable to monitor the robot's progress. P2 said, *"I think I was a bit worried about looking at the wrong area and accidentally giving the robot the wrong instructions, so I found myself checking in on the robot's progress less than with the other techniques."*

This suggests that although simple heuristics can be used to create intuitive implicit task allocation techniques, users still prefer to have additional control over how they operate through explicit mechanisms. For instance, mechanisms could allow users to control the size of the regions created by their gaze or object manipulations. Other mechanisms could allow the user to pause territory creation so that they can monitor the robot's progress or move objects around the table without unintentionally triggering new territories. This finding lends further support that mixed-initiative task allocation may be the way forward for ad-hoc HRC [2].

### 7.3 Further Enhancing HRC

The proposed allocation techniques are an initial step towards enabling more natural human-robot collaboration, but there is room for further enhancement. We think that HRC augmented by digital tabletops and mixed reality holds significant promise, and our VR simulation explored the ways in which the collaboration can be enhanced, such as with visual indicators when task allocation occurs or through visual heatmaps of territories. These visualization elements could be deployed through an overhead projector [10,35], handheld devices [7], or HMDs [43] in real-world HRC tasks. There are additional elements that could be useful for HRC tasks, such as being able to visualize the queue of objects that will be worked on by the robot. We implemented an early version of this mechanism where objects are highlighted on the table in shades of colors signifying the order in which they will be handled. It could also be beneficial to visualize which objects the robot is considering allocating when using an implicit technique. As an example, with the *distance* technique, the user could see the next few objects the robot will allocate if the objects on the table remain in their current positions. Beyond visualization aids, users could also benefit from additional automation in task allocation. For example, the robot could infer from a set of object allocations what the user might allocate next, akin to an autocomplete function. The robot could also learn users' allocation preferences during the task or upon working with them over multiple sessions. Finally, our proposed design space served as groundwork for our explored techniques but is not meant to be exhaustive. Future work could consider additional factors which could reveal new allocation techniques yet to be explored.

### 7.4 Future Work

There are many promising directions to advance this work. We opted for a VR approach partly due to the constraints of conducting in-person studies during the pandemic. However, this work was a design exploration of task allocation techniques, which benefitted from the ability to rapidly prototype and ideate beyond the constraints of today's technology. We acknowledge that some of VR's inherent limitations such as the loss of physicality from the lack of haptic feedback when grabbing objects or the low field of view may limit some of our findings or may have influenced the techniques that we proposed. Still, VR has shown promise as a prototyping and validation methodology for HRC and HRI [4,6] and future work can provide guidance on when it can be used. Conducting this study outside VR to assess if the techniques are technically feasible and to determine whether the results generalize to real-life interaction with robots could be beneficial.

An immediate extension of this work is how one might design useful explicit mechanisms for implicit allocation techniques to enhance controllability without compromising their parallelization potential. Task allocation is also meaningful in contexts where there are groups of humans and robots but would require adapting the proposed techniques which is an interesting challenge. This work assumes that the robot knows where to place objects upon picking them up, though the goal state may not be known *a priori* or be hard to specify until execution in many tasks. This leaves an open question about how the user can communicate goal information to the robot in real-time. Further, we assume that the robot is equally capable of picking up and placing objects as a human but changing this balance can alter how the techniques perform. We also noticed that users had richer interactions with the blocks than the robot. This suggests future possibilities to incorporate similar interaction capabilities to aid the robot during collaboration, such as the ability to push multiple blocks to assign them to a human. Lastly, the scope of this work was tabletop HRC but the coordination behaviors we borrow from can be equally applied to other HRC contexts such as those where teams of humans and mobile robots work together in a warehouse.

## 8 CONCLUSION

Ad-hoc HRC poses a challenge in terms of ensuring that the robot can contribute to task completion as an equal partner. We have proposed a design space of task allocation techniques that provide users with the ability to assign objects to the robot in real-time. Several of the techniques were inspired by coordination behaviors that human teams utilize in collaboration. The evaluation suggests that the techniques were generally well received, with implicit techniques lowering user burden by making the robot proactively seek task allocation, though they can be enhanced to promote greater control over robot proactivity. We hope that this work inspires significant efforts into the design of intuitive task allocation techniques for ad-hoc HRC.

## REFERENCES

[1] Till Ballendat, Nicolai Marquardt, and Saul Greenberg. 2010. Proxemic interaction: designing for a proximity and orientation-aware environment. In *ACM International Conference on Interactive Tabletops and Surfaces - ITS '10*, ACM Press, Saarbrücken, Germany, 121. DOI:https://doi.org/10.1145/1936652.1936676

[2] Jimmy Baraglia, Maya Cakmak, Yukie Nagai, Rajesh Rao, and Minoru Asada. 2016. Initiative in robot assistance during collaborative task execution. In *2016 11th ACM/IEEE International Conference on Human-Robot Interaction (HRI)*, IEEE, Christchurch, New Zealand, 67–74. DOI:https://doi.org/10.1109/HRI.2016.7451735

[3] Richard A. Bolt. 1980. Put-that-there: Voice and gesture at the graphics interface. In *Proceedings of the 7th annual conference on Computer graphics and interactive techniques* (SIGGRAPH '80), Association for Computing Machinery, New York, NY, USA, 262–270. DOI:https://doi.org/10.1145/800250.807503

[4] L. Bruckschen, K. Bungert, M. Wolter, S. Krumpen, M. Weinmann, R. Klein, and M. Bennewitz. 2020. Where Can I Help? Human-Aware Placement of Service Robots. In *2020 29th IEEE International Conference on Robot and Human Interactive Communication (RO-MAN)*, 533–538. DOI:https://doi.org/10.1109/RO-MAN47096.2020.9223331

[5] Judee K. Burgoon and Stephen B. Jones. 1976. Towards a Theory of Personal Space Expectations and Their Violations. *Human Comm Res* 2, 2 (January 1976), 131–146. DOI:https://doi.org/10.1111/j.1468-2958.1976.tb00706.x

[6] Yuanzhi Cao, Tianyi Wang, Xun Qian, Pawan S. Rao, Manav Wadhawan, Ke Huo, and Karthik Ramani. 2019. GhostAR: A Time-space Editor for Embodied Authoring of Human-Robot Collaborative Task with Augmented Reality. In *Proceedings of the 32nd Annual ACM Symposium on User Interface Software and Technology*, ACM, New Orleans LA USA, 521–534. DOI:https://doi.org/10.1145/3332165.3347902

[7] Yuanzhi Cao, Zhuangying Xu, Fan Li, Wentao Zhong, Ke Huo, and Karthik Ramani. 2019. V.Ra: An In-Situ Visual Authoring System for Robot-IoT Task Planning with Augmented Reality. In *Proceedings of the 2019 on Designing Interactive Systems Conference*, ACM, San Diego CA USA, 1059–1070. DOI:https://doi.org/10.1145/3322276.3322278

[8] Nils Dahlbäck, Arne Jönsson, and Lars Ahrenberg. 1993. Wizard of Oz studies: why and how. In *Proceedings of the 1st international conference on Intelligent user interfaces* (IUI '93), Association for Computing Machinery, New York, NY, USA, 193–200. DOI:https://doi.org/10.1145/169891.169968

[9] Jared A. Frank, Matthew Moorhead, and Vikram Kapila. 2016. Realizing mixed-reality environments with tablets for intuitive human-robot collaboration for object manipulation tasks. In *2016 25th IEEE International Symposium on Robot and Human Interactive Communication (RO-MAN)*, IEEE, New York, NY, USA, 302–307. DOI:https://doi.org/10.1109/ROMAN.2016.7745146

[10] Yuxiang Gao and Chien-Ming Huang. 2019. PATI: a projection-based augmented table-top interface for robot programming. In *Proceedings of the 24th International Conference on Intelligent User Interfaces*, ACM, Marina del Ray California, 345–355. DOI:https://doi.org/10.1145/3301275.3302326

[11] Brian Gleeson, Karon MacLean, Amir Haddadi, Elizabeth Croft, and Javier Alcazar. 2013. Gestures for industry Intuitive human-robot communication from human observation. In *2013 8th ACM/IEEE International Conference on Human-Robot Interaction (HRI)*, IEEE, Tokyo, Japan, 349–356. DOI:https://doi.org/10.1109/HRI.2013.6483609

[12] Matthew C. Gombolay, Reymundo A. Gutierrez, Shanelle G. Clarke, Giancarlo F. Sturla, and Julie A. Shah. 2015. Decision-making authority, team efficiency and human worker satisfaction in mixed human–robot teams. *Auton Robot* 39, 3 (October 2015), 293–312. DOI:https://doi.org/10.1007/s10514-015-9457-9

[13] Edward T. Hall, Ray L. Birdwhistell, Bernhard Bock, Paul Bohannan, A. Richard Diebold, Marshall Durbin, Munro S. Edmonson, J. L. Fischer, Dell Hymes, Solon T. Kimball, Weston La Barre, J. E. McClellan, Donald S. Marshall, G. B. Milner, Harvey B. Sarles, George L Trager, and Andrew P. Vayda. 1968. Proxemics [and Comments and Replies]. *Current Anthropology* 9, 2/3 (April 1968), 83–108. DOI:https://doi.org/10.1086/200975

[14] Bradley Hayes and Brian Scassellati. Challenges in Shared-Environment Human-Robot Collaboration. In *Proceedings of the "Collaborative Manipulation" Workshop at HRI 2013.*, 6.

[15] Guy Hoffman. 2019. Evaluating Fluency in Human–Robot Collaboration. *IEEE Trans. Human-Mach. Syst.* 49, 3 (June 2019), 209–218. DOI:https://doi.org/10.1109/THMS.2019.2904558

[16] Guy Hoffman and Cynthia Breazeal. 2007. Effects of anticipatory action on human-robot teamwork efficiency, fluency, and perception of team. In *Proceeding of the ACM/IEEE international conference on Human-robot interaction - HRI '07*, ACM Press, Arlington, Virginia, USA, 1. DOI:https://doi.org/10.1145/1228716.1228718

[17] Chien-Ming Huang, Maya Cakmak, and Bilge Mutlu. 2015. Adaptive Coordination Strategies for Human-Robot Handovers. In *Robotics: Science and Systems XI*, Robotics: Science and Systems Foundation. DOI:https://doi.org/10.15607/RSS.2015.XI.031

[18] Chien-Ming Huang and Bilge Mutlu. 2016. Anticipatory robot control for efficient human-robot collaboration. In *2016 11th ACM/IEEE International Conference on Human-Robot Interaction (HRI)*, IEEE, Christchurch, New Zealand, 83–90. DOI:https://doi.org/10.1109/HRI.2016.7451737

[19] Kentaro Ishii, Shengdong Zhao, Masahiko Inami, Takeo Igarashi, and Michita Imai. 2009. Designing Laser Gesture Interface for Robot Control. In *Human-Computer Interaction – INTERACT 2009*, Tom Gross, Jan Gulliksen, Paula Kotzé, Lars Oestreicher, Philippe Palanque, Raquel Oliveira Prates and Marco Winckler (eds.). Springer Berlin Heidelberg, Berlin, Heidelberg, 479–492. DOI:https://doi.org/10.1007/978-3-642-03658-3_52

[20] Lars Johannsmeier and Sami Haddadin. 2017. A Hierarchical Human-Robot Interaction-Planning Framework for Task Allocation in Collaborative Industrial Assembly Processes. *IEEE Robotics and Automation Letters* 2, 1 (January 2017), 41–48. DOI:https://doi.org/10.1109/LRA.2016.2535907

[21] Wendy Ju, Brian A. Lee, and Scott R. Klemmer. 2008. Range: exploring implicit interaction through electronic whiteboard design. In *CSCW '08: Proceedings of the 2008 ACM conference on Computer supported cooperative work*, 17–26.

[22] Ross Mead and Maja J Matarić. 2017. Autonomous human–robot proxemics: socially aware navigation based on interaction potential. *Auton Robot* 41, 5 (June 2017), 1189–1201. DOI:https://doi.org/10.1007/s10514-016-9572-2

[23] Jonathan Mumm and Bilge Mutlu. 2011. Human-robot proxemics: physical and psychological distancing in human-robot interaction. In *Proceedings of the 6th international conference on Human-robot interaction* (HRI '11), Association for Computing Machinery, New York, NY, USA, 331–338. DOI:https://doi.org/10.1145/1957656.1957786

[24] Camilo Perez Quintero, Romeo Tatsambon Fomena, Azad Shademan, Nina Wolleb, Travis Dick, and Martin Jagersand. 2013. SEPO: Selecting by pointing as an intuitive human-robot command interface. In *2013 IEEE International Conference on Robotics and Automation*, IEEE, Karlsruhe, Germany, 1166–1171. DOI:https://doi.org/10.1109/ICRA.2013.6630719

[25] Camilo Perez Quintero, Romeo Tatsambon, Mona Gridseth, and Martin Jägersand. 2015. Visual pointing gestures for bi-directional human robot interaction in a pick-and-place task. In *2015 24th IEEE International Symposium on Robot and Human Interactive Communication (RO-MAN)*, 349–354. DOI:https://doi.org/10.1109/ROMAN.2015.7333604

[26] Fabian Ranz, Vera Hummel, and Wilfried Sihn. 2017. Capability-based Task Allocation in Human-robot Collaboration. *Procedia Manufacturing* 9, (January 2017), 182–189. DOI:https://doi.org/10.1016/j.promfg.2017.04.011

[27] Alessandro Roncone, Olivier Mangin, and Brian Scassellati. 2017. Transparent role assignment and task allocation in human robot collaboration. In *2017 IEEE International Conference on Robotics and Automation (ICRA)*, IEEE, Singapore, Singapore, 1014–1021. DOI:https://doi.org/10.1109/ICRA.2017.7989122

[28] Daisuke Sakamoto, Koichiro Honda, Masahiko Inami, and Takeo Igarashi. 2009. Sketch and run: a stroke-based interface for home robots. In *Proceedings of the 27th international conference on Human factors in computing systems - CHI 09*, ACM Press, Boston, MA, USA, 197. DOI:https://doi.org/10.1145/1518701.1518733

[29] K. Sakita, K. Ogawam, S. Murakami, K. Kawamura, and K. Ikeuchi. 2004. Flexible cooperation between human and robot by interpreting human intention from gaze information. In *2004 IEEE/RSJ International Conference on Intelligent Robots and Systems (IROS) (IEEE Cat. No.04CH37566)*, IEEE, Sendai, Japan, 846–851. DOI:https://doi.org/10.1109/IROS.2004.1389458

[30] Allison Sauppé and Bilge Mutlu. 2014. Robot deictics: how gesture and context shape referential communication. In *Proceedings of the 2014 ACM/IEEE international conference on Human-robot interaction - HRI '14*, ACM Press, Bielefeld, Germany, 342–349. DOI:https://doi.org/10.1145/2559636.2559657

[31] Ruth Schulz, Philipp Kratzer, and Marc Toussaint. 2018. Preferred Interaction Styles for Human-Robot Collaboration Vary Over Tasks With Different Action Types. *Front. Neurorobot.* 12, (2018). DOI:https://doi.org/10.3389/fnbot.2018.00036

[32] Stacey D. Scott and Sheelagh Carpendale. 2010. Theory of Tabletop Territoriality. In *Tabletops - Horizontal Interactive Displays*, Christian Müller-Tomfelde (ed.). Springer London, London, 357–385. DOI:https://doi.org/10.1007/978-1-84996-113-4_15

[33] Stacey D. Scott, M. Sheelagh, T. Carpendale, and Kori M. Inkpen. 2004. Territoriality in collaborative tabletop workspaces. In *Proceedings of the 2004 ACM conference on Computer supported cooperative work - CSCW '04*, ACM Press, Chicago, Illinois, USA, 294. DOI:https://doi.org/10.1145/1031607.1031655

[34] Yasaman S. Sefidgar and Maya Cakmak. 2018. End-User Programming of Manipulator Robots in Situated Tangible Programming Paradigm. In *Companion of the 2018 ACM/IEEE International Conference on Human-Robot Interaction*, ACM, Chicago IL USA, 319–320. DOI:https://doi.org/10.1145/3173386.3176923

[35] Yasaman S. Sefidgar, Thomas Weng, Heather Harvey, Sarah Elliott, and Maya Cakmak. 2018. RobotIST: Interactive Situated Tangible Robot Programming. In *Proceedings of the Symposium on Spatial User Interaction* (SUI '18), Association for Computing Machinery, New York, NY, USA, 141–149. DOI:https://doi.org/10.1145/3267782.3267921

[36] Julie Shah and Cynthia Breazeal. 2010. An Empirical Analysis of Team Coordination Behaviors and Action Planning With Application to Human–Robot Teaming. *Hum Factors* 52, 2 (April 2010), 234–245. DOI:https://doi.org/10.1177/0018720809350882

[37] Julie Shah, James Wiken, Brian Williams, and Cynthia Breazeal. 2011. Improved human-robot team performance using chaski, a human-inspired plan execution system. In *Proceedings of the 6th international conference on Human-robot interaction - HRI '11*, ACM Press, Lausanne, Switzerland, 29. DOI:https://doi.org/10.1145/1957656.1957668

[38] Robert Sommer. 1969. Personal Space. The Behavioral Basis of Design. (1969).

[39] Aaron St. Clair and Maja Matric. 2015. How Robot Verbal Feedback Can Improve Team Performance in Human-Robot Task Collaborations. In *Proceedings of the Tenth Annual ACM/IEEE International Conference on Human-Robot Interaction* (HRI '15), Association for Computing Machinery, New York, NY, USA, 213–220. DOI:https://doi.org/10.1145/2696454.2696491

[40] Panagiota Tsarouchi, Alexandros-Stereos Matthaiakis, Sotiris Makris, and George Chryssolouris. 2017. On a human-robot collaboration in an assembly cell. *International Journal of Computer Integrated Manufacturing* 30, 6 (June 2017), 580–589. DOI:https://doi.org/10.1080/0951192X.2016.1187297

[41] Daniel Vogel and Ravin Balakrishnan. 2004. Interactive Public Ambient Displays: Transitioning from Implicit to Explicit, Public to Personal, Interaction with Multiple Users. In *Proceedings of the 17th annual ACM symposium on User interface software and technology*, 137–146.

[42] Michael E. Walker, Hooman Hedayati, and Daniel Szafir. 2019. Robot Teleoperation with Augmented Reality Virtual Surrogates. In *2019 14th ACM/IEEE International Conference on Human-Robot Interaction (HRI)*, IEEE, Daegu, Korea (South), 202–210. DOI:https://doi.org/10.1109/HRI.2019.8673306

[43] Michael Walker, Hooman Hedayati, Jennifer Lee, and Daniel Szafir. 2018. Communicating Robot Motion Intent with Augmented Reality. In *Proceedings of the 2018 ACM/IEEE International Conference on Human-Robot Interaction*, ACM, Chicago IL USA, 316–324. DOI:https://doi.org/10.1145/3171221.3171253